\documentclass[10pt,a4paper]{article}   

\usepackage{amsmath,,amssymb,mathrsfs,bm,enumerate,physics}
\usepackage{autobreak,mathtools}
\usepackage{tikz,float,placeins}
\usetikzlibrary{intersections,calc,arrows.meta}
\usepackage{ulem}
\usepackage{hyperref}
\usepackage{cleveref}
\usepackage{graphicx,booktabs,capt-of}
\usepackage{chngcntr}

\usepackage{spverbatim}
\usepackage{caption}
\usepackage{comment}
\usepackage{tabularx}
\usepackage{enumitem}
\usepackage{indentfirst}
\usepackage{changepage}

\newcommand\myshade{75}
\colorlet{mylinkcolor}{red}
\colorlet{mycitecolor}{green}
\colorlet{myurlcolor}{blue}
\hypersetup{colorlinks=true, anchorcolor=cyan, 
linkcolor=mylinkcolor!\myshade!black, 
citecolor=mycitecolor!\myshade!black,
urlcolor=myurlcolor!\myshade!black } 
\newcommand{\beq}{\begin{equation}}
\newcommand{\eeq}{\end{equation}}

\newcommand{\ra}{\rightarrow}

\makeatletter
\long\def\@makefntext#1{\parindent 1em\noindent
\@hangfrom{\hbox to 1.8em{\hss$^{\@thefnmark}$}}#1}
\makeatother

\usepackage[top=20mm, bottom=20mm, left=25mm, right=25mm]{geometry}

\begin{document}

\title{Ensemble Bayesian Inference: Leveraging Small Language Models to Achieve LLM-level Accuracy in Profile Matching Tasks}

\author{Haru-Tada Sato,  Fuka Matsuzaki and Jun-ichiro Takahashi \\
 Department of Data Science, i's Factory Corporation, Ltd.\\
       Kanda-nishiki-cho 2-7-6, Tokyo 101-0054, JAPAN\\
{\tt \{satoh, matsuzaki, jtakahashi\}@isfactory.co.jp }}
\date{}

\maketitle
\begin{abstract}
This study explores the potential of small language model (SLM) ensembles to achieve accuracy comparable to proprietary large language models (LLMs). We propose Ensemble Bayesian Inference (EBI), a novel approach that applies Bayesian estimation to combine judgments from multiple SLMs, allowing them to exceed the performance limitations of individual models. Our experiments on diverse tasks—aptitude assessments and consumer profile analysis in both Japanese and English—demonstrate EBI's effectiveness. Notably, we analyze cases where incorporating models with negative Lift values into ensembles improves overall performance, and we examine the method's efficacy across different languages. These findings suggest new possibilities for constructing high-performance AI systems with limited computational resources and for effectively utilizing models with individually lower performance. Building on existing research on LLM performance evaluation, ensemble methods, and open-source LLM utilization, we discuss the novelty and significance of our approach.~\footnote{Codes and data are available at 
https://github.com/maskcode9004/ebi\_method}
\end{abstract}

Ketwords:
Bayesian Inference, Ensemble Learning, Profiling Tasks, Proprietary LLM Comparison, Small Language Models

\setlength{\parindent}{1em}
\section{Introduction}

In the field of medical diagnosis, there have been reports of GPT-4 achieving higher diagnostic accuracy than physicians, either independently or as an assistant~\cite{ref1,ref2}, showing comparable performance~\cite{ref3}, or failing to demonstrate improvement~\cite{ref4}. While some research emphasizes the irreplaceability of specialized knowledge and experience and shows caution about AI's ability to accurately identify common sense in problem-solving, studies also suggest its effectiveness as a supportive tool~\cite{ref5,ref6}. Various diagnostic tasks have been evaluated, including predictive accuracy from test results and case studies~\cite{ref1,ref3}, as well as diagnostic summary content~\cite{ref2}.

In the field of psychology, comparative experiments are being conducted to determine whether AI can substitute human intellectual activities such as decision-making and cognition~\cite{ref7}-~\cite{ref11}. In creative ideation, AI demonstrates novelty equivalent to or surpassing humans, though it shows slight deficiencies in feasibility~\cite{ref7}. Psychological experiments evaluate whether AI can replace human subjects~\cite{ref8} or make human-like behavioral judgments~\cite{ref9}. Experiments involving human judgment present challenges for quantitative evaluation, including issues of fairness due to human cognitive biases~\cite{ref10}, handling subjective concepts in evaluations~\cite{ref11}, responsiveness to cognitive load~\cite{ref12}, and variability in assessments~\cite{ref13}.

Against this background, this paper addresses person identification as a judgment task in comparative experiments between AI and humans. The judgment targets are texts evaluating personal characteristics. The task involves comparing two types of evaluation comments about the same group of individuals and identifying each person, of which samples are illustrated in Appendix~\ref{sec:ABtxt}. Evaluation texts contain noisy expressions or potentially misleading statements dependent on the evaluator, posing risks of misjudgment. Improving the robustness and reliability of inference when faced with such inputs presents a significant challenge.

To execute such complex and sophisticated judgments using AI, we propose a judgment methodology that combines probabilistic judgment using Bayesian Inference (BI) with subjective elements based on confidence levels, taking into account human judgment characteristics (variability and pattern consistency~\cite{ref13}). While the BI method is a simplified model of human judgment processes, generative AI also exhibits judgment variability, which we aim to accommodate through ensemble approach to BI. This is the 
key idea that we call our method Ensemble Beyesian Inference (EBI).

While ensemble methods for generative AI~\cite{ref14}-~\cite{ref17} focus on improving model accuracy, this paper examines whether judgment diversity contributes to accuracy improvement, centering on recently emerged lightweight open-source LLMs. As addressing computational costs and latency associated with advanced prompting techniques becomes crucial — especially when targeting real-world applications with stringent performance requirements — our work also serves to validate a framework for high-volume processing leveraging the advantages of high-speed processing chips to bridge the performance gap with proprietary LLMs.

%
%
\section{Task Design of Bayesian Inference Method}

The task problem setting involves determining which element $A(a_j)$ corresponds to the same individual when given all information $X=B(b_i)$, with two lists of personal information $B=\{b_1,b_2, \cdots, b_n\}$ and $A=\{a_1,a_2, \cdots, a_n\}$ about a group of individuals. Here, $A$ and $B$ represent sets of texts describing personal profiles based on different perspectives. By "different perspectives," we mean domains that have some relationship, such as dietary habits and health hygiene.

The goal is to estimate the profiles of each person and determine whether they are the same individual through profile analysis. Technically, the main objective is to achieve judgment accuracy comparable to proprietary LLMs by obtaining frequent responses from SLMs and implementing a probabilistic judgment process. The extent to which performance can exceed human judgment is also a significant interest.

\subsection{Confidence Matrix and Subjective Degree}

The basic concept of our technique is the Bayes' theorem. The conditional probability of event A occurring after event X (posterior probability) $P(A|X)$ is given by
likelihood $P(X|A)$ and $P(A)$, the probability of event A occurring (prior probability or subjective probability):
\beq
P(A|X)=\frac{P(X|A)P(A)}{P(X)}\,, \label{bayes}
\eeq
where $P(X)=\sum_A P(X|A)P(A)$.


When an element $b_i$ of $B=\{b_1,b_2, \cdots, b_n\}$ is given, the posterior probability that it corresponds to an element $a_j$ of $A=\{a_1,a_2, \cdots, a_n\}$ provides the confidence matrix $\mathrm{conf}_{B\ra A}$ (statistical or computational confidence, simply referred to as confidence when distinction is unnecessary).
Its elements, with $X=B(b_i)$, are given by Bayes' theorem \eqref{bayes}:
\beq
(\mathrm{conf}_{B\ra A})_{ij}=
P(a_j|b_i)=\frac{P(b_i|a_j)P(a_j)}{P(b_i)}\,. \label{confmat}
\eeq

An experimental consideration is that when aggregating responses from multiple queries to calculate confidence, elements that never appear in responses would result in $P(X)=0$, potentially leading to divergent matrix elements in naive aggregation. Therefore, we apply regularization to ensure $P(X)$ has a small value $\varepsilon$. Since $P(X|A)$ values are also proportionally small, equation \eqref{bayes} produces finite values. Specifically, items with no responses are assigned a value of $\varepsilon=0.1$ in aggregation to avoid division by zero.


Conversely, when judging $B(b_i)$ from $A(a_j)$, we obtain the strength of confidence as an observed value. To distinguish this from the above-mentioned confidence, we call it subjective confidence (subjective degree). In a voting system, this could be the proportion of votes received (collective subjective degree). When this subjective degree is observed as $c_{ji}$, the observation matrix (subjective degree matrix), 
the likelihood $P(b_i|a_j)$ is given by $(\mathrm{conf}_{A\ra B})_{ij} = c_{ji}$, $(0\leq c_{ji} \leq1)$. 
Then we obtain the subjective probability: 
$P(a_j)=\frac{1}{C}\sum_i c_{ji}$, where $C=\sum_{ij} c_{ji}$,
yielding
\beq
P(b_i)=\sum_j P(b_i|a_j)P(a_j)=\frac{1}{C}\sum_{jk} c_{ji}c_{jk}\,.
\eeq
Substituting these into \eqref{confmat} gives the confidence matrix.

\subsection{Judgment Matrix}

The problem setting is to determine which $a_j$ corresponds to $b_i$ when given information $X=b_i$. Simple Bayesian estimation may not work well when there is information asymmetry.Namely, probabilistic LLM  pproaches, such as MLM and GPT, may not perform reliably when profiler differences in skill, perspective, or background affect interpretation of $A$ and $B$.

Regardless of the method used, if the reliability of the judgment result can be expressed through some evaluation weight $s_{ij}$, it appears effective to define a judgment matrix $J$ by multiplying this by the confidence degree:
\beq
J_{ij}=s_{ij} P(a_j|b_i) \,.
\eeq
The $ij$ component of $J$ can be interpreted as the plausibility of each candidate $a_j$ for $b_i$, so this $J$ is used for identity judgment. This method allows for various models to be constructed based on the freedom in defining $s_{ij}$ and the observation matrix $c_{ji}$.

\subsection{System Settings}

In this study, we use small-size language models (SLMs) to determine whether $b_i$ and $a_j$ represent the same person to create $s_{ij}$. However to expand the range of system options, we test multiple different LMs for calculating $s_{ij}$ and $c_{ij}$. 
For obtaining observation matrix $c_{ij}$ and weight matrix $s_{ij}$, we consider two approaches: (i) evaluation based on collective subjective degree (id selection frequency) and (ii) evaluation based on direct subjective degree. Thease are realized by the following 
types of prompt processing: 
\begin{quote}
Type 1. Aggregating the frequency of ids obtained as responses by submitting the same question multiple times\par
Type 2. Collecting values by askig for subjective confidence levels during the response evaluation process
\end{quote}
\vspace{-1em}

The prompt structure varies depending on the Type, but the basic query asking which id is identical and the specification of the analysis method are common across both. What's common is the concept, not necessarily the specific descriptive expressions. The essential difference lies in the output format specification: Type 1 outputs only the pairs of ids judged identical (see Appendix \ref{t1}), while Type 2 outputs an array of candidate $a_j$ in descending order of subjective degree along with their subjective degrees (see Appendix \ref{t2}).
These aggregate values are normalized by the number of questions.

In calculating observation and weight matrices, responses will differ depending on which language model is used and what prompt (Type and descriptive expression) is given, so we consider one {\it language model and prompt configuration} as one system. Additionally, we consider ensemble systems (EBI) by calculating weighted averages of judgment matrices $J$ computed from individual systems.

For language model selection criteria, we primarily adopt lightweight SLMs that prioritize speed over accuracy for collecting multiple responses, however we also consider 70b models for some cases using high-speed processing via WSE (Cerebras Wafer-Scale Engine and CS-3 system) or LPU (Groq LPU™ Inference Engine). Models used in EBI include:
GroqChat: gemma2-9b-it, mixtral-8x7b-32768, llama3-8b-8192, llama3-70b-8192.
Cerebras: llama-3.1-70b-versatile. OpenAI: gpt-4o-mini-2024-07-18.

%
%
\section{Data Generation and Evaluation Methods}

\subsection{Data Preparation}
We generate two different persona profiles for the same individuals based on items (attribute values and observed values) from multiple perspectives. We divide the items used for generation into two parts, allowing overlap, and create two types of persona data, $A$ and $B$. When creating these personas, we specify different perspectives to ensure they cannot be easily linked merely by word matching (similarity). We prepared two types of datasets: aptitude assessments and purchase history. For sample data, refer to Appendix~\ref{sec:ABtxt}.

For the aptitude assessment (prof1j), the specific creation process is as follows: The observational data consists of 50 people's aptitude test results with 14 observation items (items including sociability avoidance, self-reflection, task persistence, risk avoidance level, etc.). Character diagnosis comments were generated using GPT-4o based on the item names and numerical values to create dataset $A$. Department, years of service, and other work attributes were added to similarly generate work evaluation comments to create dataset $B$. In creating the comments, different perspectives were specified to ensure that $A$ and $B$ would not generate identical comments. For $A$, the purpose was to encourage self-improvement to enhance performance, while $B$ focused on issues related to work execution. Additionally, another dataset "prof1e" was created by translating this data to English using GPT-4o.

For the purchase history data (prof2j), 18 purchased product items were divided, and persona data was created using a similar procedure as above. In this case, since the persona image could be significantly affected by the product composition in the two divisions, both $A$ and $B$ were created with two types of perspectives: behavioral patterns and consumer values in product selection. An English translation of this data was created as "prof2e".

\subsection{Evaluation Methods}
For performance evaluation metrics, accuracy Acc$=n_c/N$ (data count $N=50$, correct answer count $n_c$) is commonly used, but if the background contexts of datasets $A$ and $B$ differ, the background knowledge of words and the range of conceptual similarities may also differ between them. In such situations with conceptual fluctuations and undefined ranges, no known method objectively quantifies the consistency of meta-concepts. Thus, a key consideration is that the scale can vary significantly due to the nature of text content, abstraction level, interpretability, and the evaluator's background knowledge, making comparison across different datasets impossible. Therefore, this metric can only be used for relative superiority judgments within the same type of data, which is a disadvantage.

Consequently, evaluating using improvement rate (Lift), which shows how much superiority there is relative to human judgment, allows for more appropriate comparison by somewhat eliminating data dependence (although the prerequisite is that the reference evaluator should remain fixed). Given the maximum correct answer count $H$ of the reference evaluator, Lift$=100(\frac{n_c}{H}-1)$ [\%].

When human evaluators are difficult to secure, high-precision proprietary LLMs can be substituted for human evaluators. In this case, using the maximum correct answer count $G$ of the high-precision proprietary LLM (ChatGPT-4o), we evaluate how close we can get using the reach rate Reach$=100n_c/\mbox{Base}$ [\%] (where Base is $H$ for humans and $G$ for LLMs). For the reference evaluator LLM, rather than using the EBI under examination, we adopt a judgment method that builds reasoning step by step through prompting to approach human judgment as closely as possible (see Section \ref{sec:LLM}).

For our English verification where $H$ is unknown, we use the average H/G ratio $\gamma$ from the Japanese verification, substituting $H=G\gamma$ into the Lift formula to approximate Lift as $\mbox{Lift}_{eff}=\mbox{Reach}/\gamma-100$. We obtained $H$ values of (19, 13) for (prof1j, prof2j) from human evaluations. For the four datasets (prof1j, prof2j, prof1e, prof2e), the obtained $G$ values were (22, 20, 28, 23). Therefore, the H/G ratios were 19/22 for prof1j and 13/20 for prof2j, giving $\gamma=0.757$. For English data, we use $\mbox{H}_{eff}=\mbox{G}\gamma$, thus adopting baseline $H$ values of (19, 13, 21.2, 17.4) for the four datasets. Human Acc values were (38\%, 26\%, 42\%, 35\%), GPT-4o Acc values were (44\%, 40\%, 56\%, 46\%), and Lift values relative to humans were (15.8\%, 53.8\%, 32.1\%, 32.2\%).

%
%
\section{Sequential Thinking Method}\label{sec:LLM}

Even proprietary LLMs require dynamic prompting techniques — where language models improve task performance through self-reflection — to ensure at least human-level evaluation accuracy \cite{ref15,ref18,ref19}.
Since simple task instructions can result in duplicate judgments from LLMs, this paper applies the Feedback-Reflect-Refine mechanism \cite{ref15} to resolve conflicts by collecting error information from LLM outputs and using it as feedback.

The prompting uses two types of templates for solution and verification:
(i) Solving Prompt: 
Instructs the model to explain the reason in one sentence before providing the answer corresponding to the task.
(ii) Feedback Prompt: 
Analyzes the cause when the model's output is incorrect and generates a new prompt.

In contrast, the EBI method eliminates duplications through forceful processing rather than prompt-based judgment. Specifically, it follows the steps:
1. Select the element with the highest output value in the judgment matrix $J$.
2. Remove the row and column containing the selected element (i.e., form a submatrix).
3. Select the element with the highest output value in the submatrix.
4. Repeat the above steps for the number of rows in the judgment matrix (i.e., the number of individuals to be predicted).

\subsection{Digest of Sequential Thinking}
In the sequential method, candidates are analyzed one by one within subsets filtered by attributes such as age (system prompt S1), and candidates are narrowed down tournament-style (sequential processing S2). Since an incorrectly selected initial candidate affects subsequent judgments, a flow to review the results (recursive processing S3) is incorporated when there are conflicts or inconsistencies in the judgment results.
Since conflicts may also occur during the review process, examination and judgment are repeated until conflicts are resolved (conflict processing S4). Refer to Appendix~\ref{sec:Spr} for concrete prompts S1-S4.
The sequential processing outline is as follows:
\begin{itemize}
\item[1.] The program obtains the set of ids from A (Aset) that match demographic attributes such as age. The number of elements is denoted as $n$.
  When Aset changes, the message history is cleared (session switching)
\item[2.] System prompt S1: Specifies the data description and analysis task method.
\item[3.] Sequential processing S2: If $n=1$, the identification is confirmed, and the process moves to the next Aset. If $n\geq2$, starting with an initial value $k:=n$, elements are removed from Aset with each identification until $k=1$.
\item[4.] When the remaining number of people in Aset reaches a specified value, recursive processing S3 is activated.
\item[5.] Conflict resolution S4 is executed within the recursive processing. The loop continues until the duplicate counter reaches 0.
\end{itemize}

%
%
\section{Results}

In this study, we conducted experiments on four datasets to examine whether SLM ensemble judgment systems can achieve judgment accuracy comparable to human or proprietary LLM judgment. The complete results of BI and EBI methods for 
each system are listed in Appendix~\ref{sec:alldata}.

\subsection{Overall}
\begin{figure}[h]	
	\begin{minipage}{1.0\hsize}
	\centering
        \captionsetup{width=.85\linewidth}
        \includegraphics[width=0.45\linewidth]{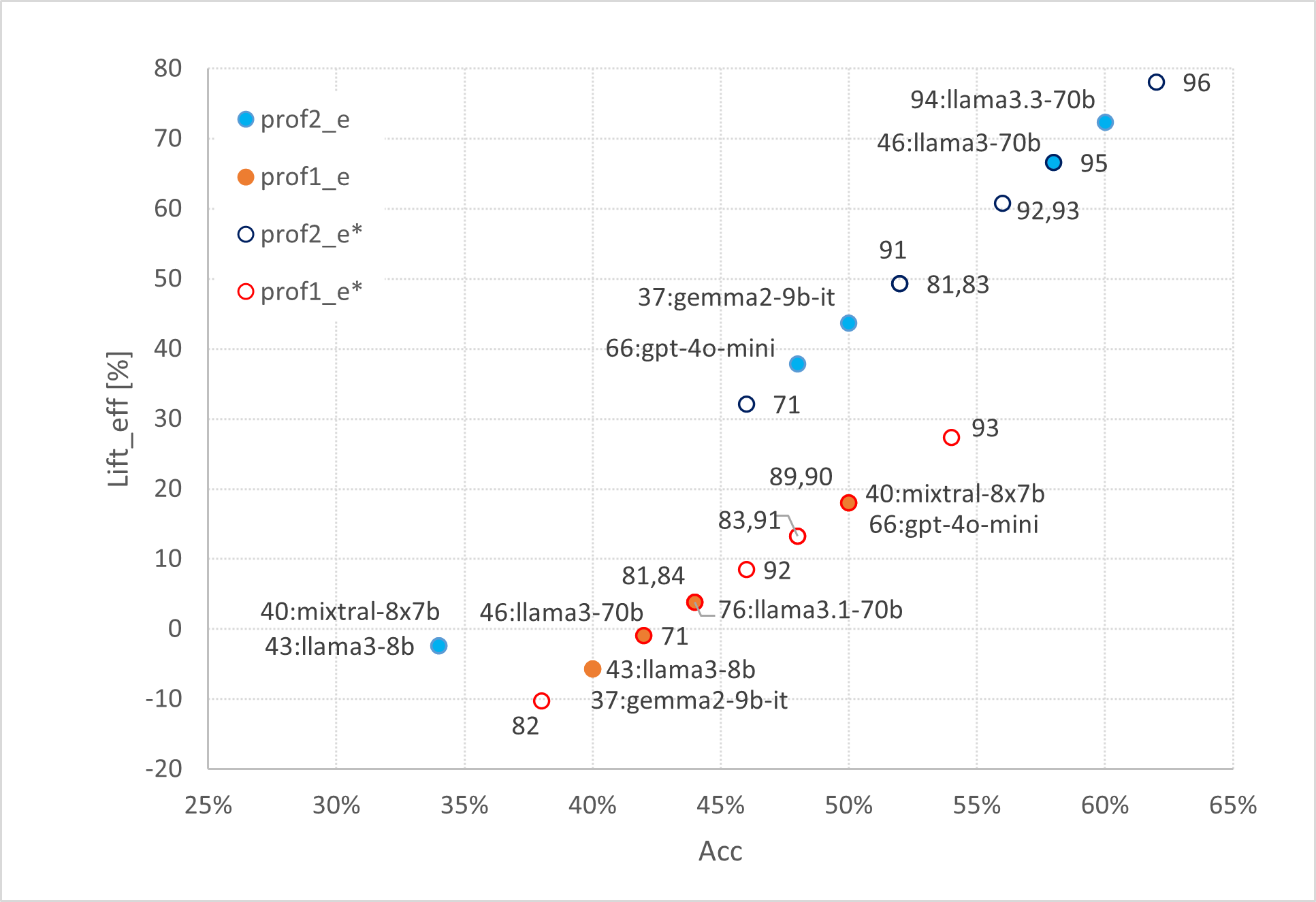}
        \includegraphics[width=0.45\linewidth]{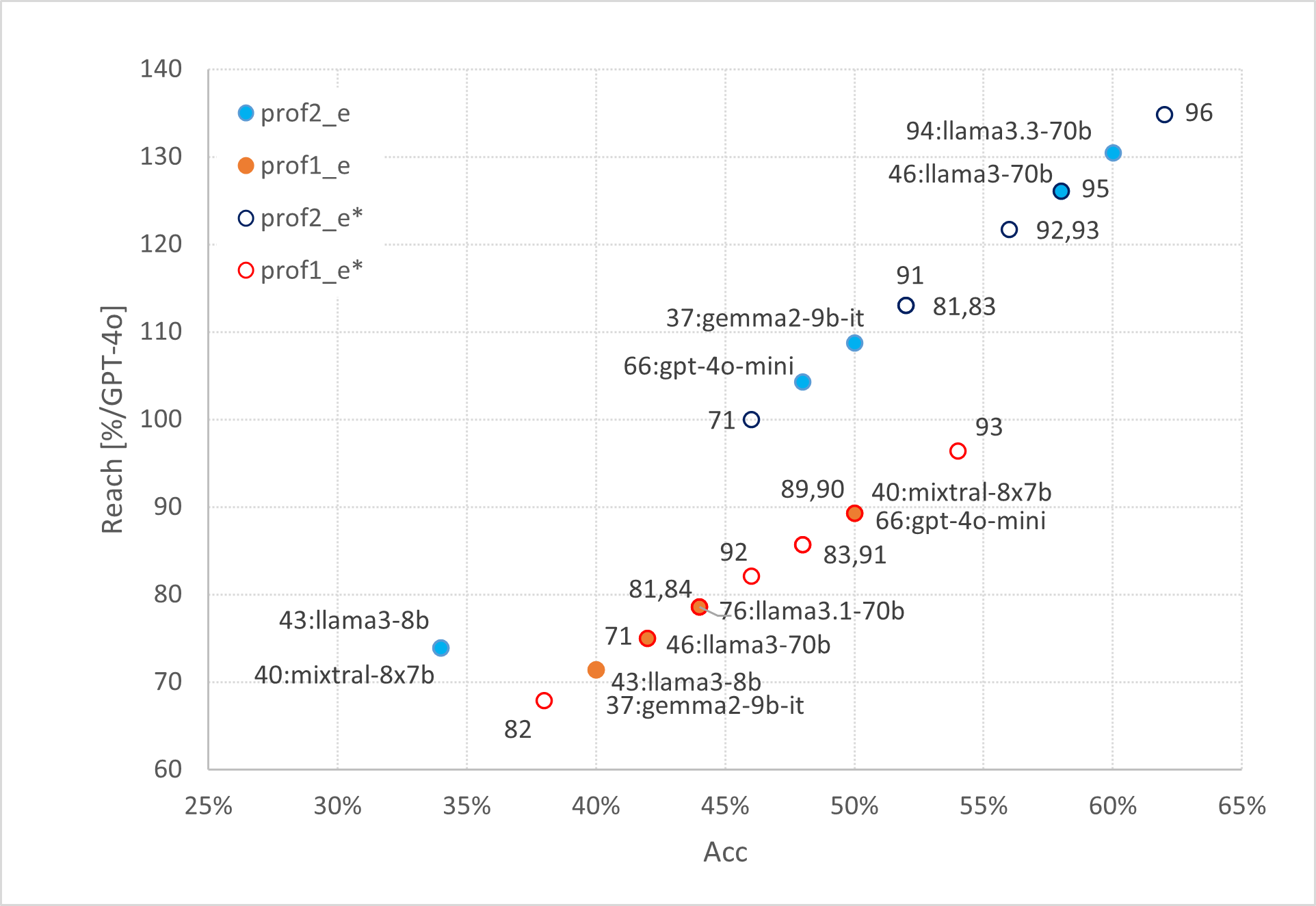}
        \vspace{-0.1cm}
    	\caption{Results of English datasets. The Lift (left) and the Reach (right) graphs for single BI systems (prof1\_e and prof2\_e) marked in filled cirlces, and EBI systems (prof1\_e* and prof2\_e*) marked in hollow circles. The component systems of EBI are listed in Appendices~\ref{sec:prof1e} and \ref{sec:prof2e}.}
     \label{fig:results_e}
     \vspace{-0.35cm}
	\end{minipage}
\end{figure}

Figure~\ref{fig:results_j} summarizes the results for Japanese datasets, and Figure~\ref{fig:results_e} summarizes the results for English datasets. Considering all four types of datasets collectively, we found that (1) single BI systems can achieve Lift$>$0, and (2) using EBI enables us to obtain systems with even higher Lift than single BI systems. In each Figure, considering Lift$\gtrsim$30\%, Reach$\gtrsim$100\% as high-performance regions, all datasets except prof1e (aptitude/English) reached these regions, and even prof1e nearly reached the lower bound of this region when using EBI (hollow markers show higher Lift values than filled markers). With the exception of system 59 (llama3.1-70b) in prof2j (purchase/Japanese), we were able to achieve higher Acc with EBI systems (compared to single BI) for all datasets. It should be noted, however, that since the number of plots can be increased indefinitely by increasing the number of models and systems tested, there is no fair way to compare the averages of Acc or Lift, so it is appropriate to understand these as trends.

%
%
\begin{figure}[h]	
	\begin{minipage}{1.0\hsize}
	\centering
        \captionsetup{width=.85\linewidth}
        \includegraphics[width=0.45\linewidth]{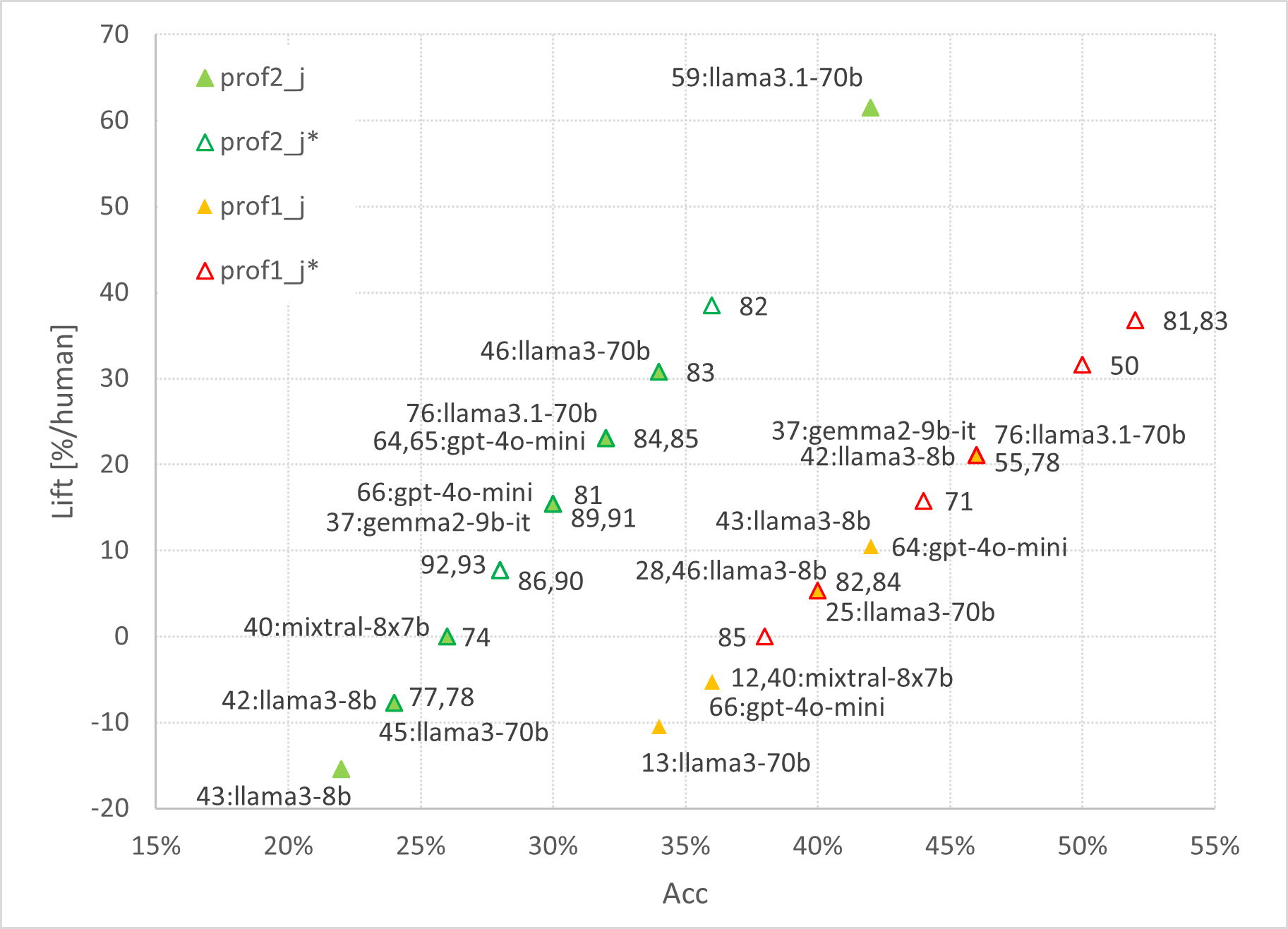}
        \includegraphics[width=0.45\linewidth]{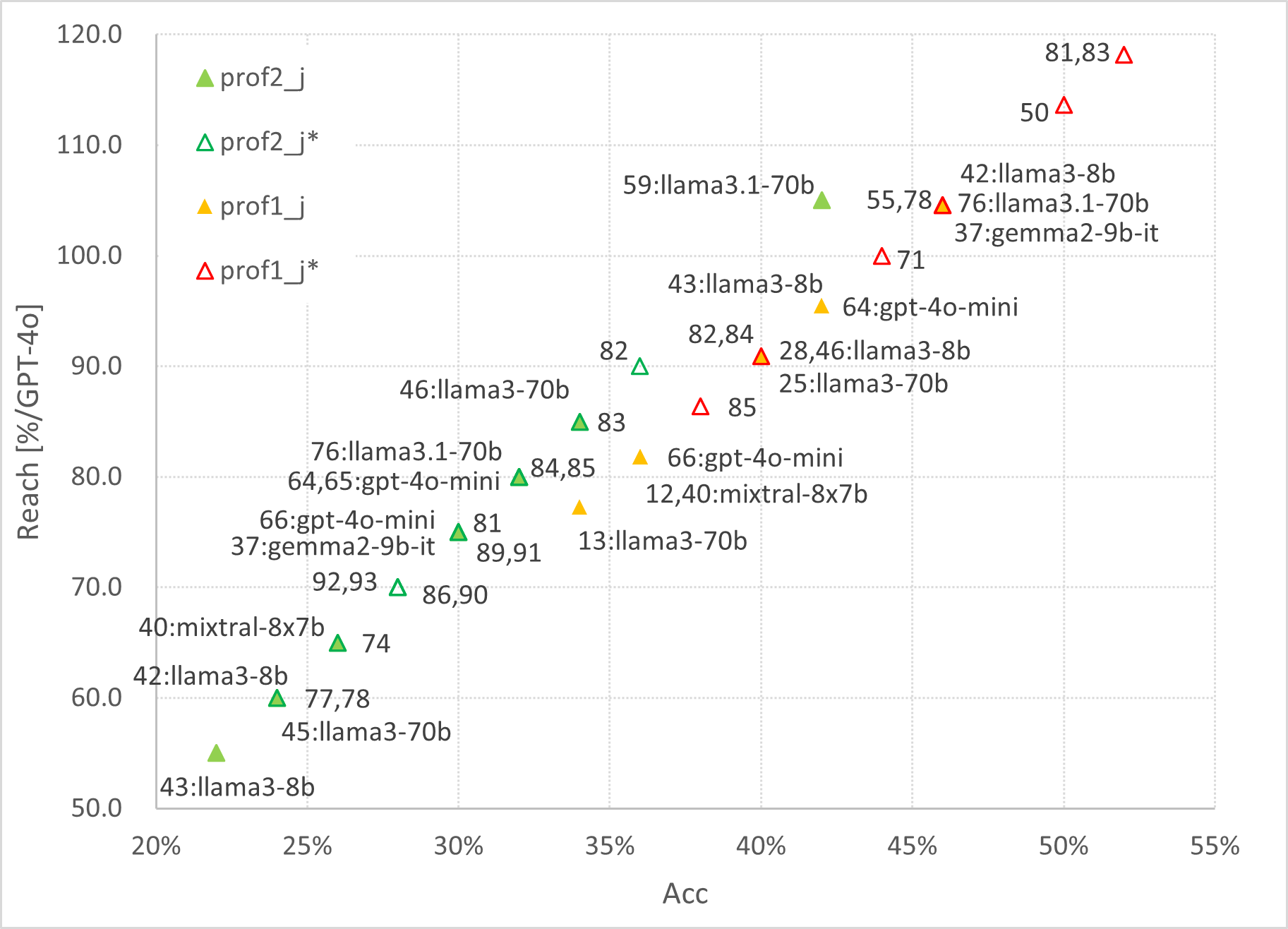}
        \vspace{-0.1cm}
    	\caption{Results of Japanses datasets. The Lift (left) and the Reach (right) graphs for single BI systems (prof1\_j and prof2\_j) marked in filled triangles, and EBI systems (prof1\_j* and prof2\_j*) marked in hollow triangles. The component systems of EBI are listed in Appendices~\ref{sec:prof1j} and \ref{sec:prof2j}. }
     \label{fig:results_j}
     \vspace{-0.35cm}
	\end{minipage}
\end{figure}

%
%

\subsection{Individual Datasets}
These results suggest that SLM ensembles using the EBI method can exceed the performance limitations of individual BI systems and achieve accuracy comparable to, or in some cases surpassing, that of LLMs. Furthermore, examining the performance of individual BI systems constituting the ensembles revealed that combining lower-performing systems appropriately can enhance prediction robustness. The confirmation of ensemble effects across different datasets and languages is an important finding demonstrating the versatility of this method. Specifically:

For prof1j (aptitude/Japanese), evaluation of multiple SLMs' individual performance revealed systems achieving maximum Lift values of 21\% (37:gemma2-9b-it, 42:llama3-8b-8192, 76:llama3.1-70b-versatile). Details are in Appendix~\ref{sec:prof1j}. By constructing average systems including these high-performing single systems, the (Lift,Reach) values (hereinafter, LR-value) increased significantly from single systems to a maximum of (36.8\%,118.2\%). Notably, in higher accuracy ensemble systems (limited to Lift$>0$), instances were confirmed where combining single systems with negative Lift values (66:gpt-4o-mini-2024-07-18, 12 and 40:mixtral-8x7b-32768, 13:llama3-70b-8192, etc.) resulted in positive Lift values after ensemble. This suggests that EBI ensembles can compensate for weaknesses in individually low-performing systems, enabling more robust predictions. For example, the ensemble system shown in Table~\ref{tb:complex_pro1} achieves positive Lift values despite including components with negative Lift.

In the prof2j (purchase/Japanese) experiment summarized in Appendix~\ref{sec:prof2j}, ensemble systems were constructed centered on systems (37, 46, 64, 76) that achieved positive Lift values acrosss two different Japanese datasets. Although Reach did not reach 100\%, instances were confirmed where including weak systems (40, 42, 43) with negative or zero Lift values in the ensemble led to positive Lift values, here too. In particular, the top-performing system 82 (in Table~\ref{tb:complex_pro2}) includes llama3-70b systems 46 and 76, and by integrating them as an average system, it achieved results (38.5\%,90\%) surpassing the maximum LR-value of each component system (30.8\%,85\%).

In the prof1e (aptitude/English) in Appendix~\ref{sec:prof1e}, although the maximum LR-value of ensemble systems (27.4\%,96.4\%) did not reach 100\% Reach, it achieved significantly higher LR-values than single systems. Again, performance improvements from single systems were observed through ensemble effects with the combination of 37 (gemma2-9bit), 40 (mixtral-8x7b-32768), 43 (llama3-8b-8192), and 76 (llama3.1-70b-versatile), with 37 and 43 proving effective as weak learners with Lift values$<0$ as understood from Table~\ref{tb:single_pro1e}.

In the prof2e (purchase/English) case (Appendix~\ref{sec:prof2e}), Table~\ref{tb:complex_pro2e} shows that ensemble systems combining multiple systems achieved significantly higher LR-values than single systems, with a maximum of (78.2\%,134.8\%). Specifically, notable performance improvements were observed by combining systems 37 (gemma2-9b-it), 40 (mixtral-8x7b-32768), 43 (llama3-8b-8192), 46 (llama3-70b-8192), 66 (gpt-4o-mini-2024-07-18), etc., with various weightings. Systems 40 and 43 functioned as weak learners with Lift values$<0$ as seen in  Table~\ref{tb:single_pro2e}.

%
%
\section{Conclusions}

From the experimental results obtained in this research, the following conclusions can be drawn:
1. Enhanced Performance through SLM Ensembles: Using ensemble techniques such as EBI makes it possible to combine multiple lightweight SLMs to exceed the performance limitations of individual systems and achieve accuracy comparable to LLMs. This suggests the possibility of constructing high-performance NLP systems in environments with limited computational resources.

2. Utilization of Weak Learners: Instances were confirmed where incorporating systems showing negative Lift values, i.e., individually low-performing systems, into ensembles improved overall performance. This reevaluates the role of "weak learners" in ensemble learning and demonstrates the effectiveness of integrating multiple systems with diverse perspectives.

3. Effectiveness of the EBI Method: The EBI method, which applies weighting using collective subjective degrees or direct subjective degrees, was suggested to be more effective at integrating individual system judgments and improving ensemble performance compared to simple averaging or other methods.

4. Versatility of the Method: Performance improvements through SLM ensembles were confirmed across different tasks such as aptitude assessment and consumer analysis, as well as across different languages including Japanese and English. This suggests that the proposed method is not limited to specific tasks or languages but has broad applicability.

However, several challenges were also identified in this research. Since the metrics are based on comparisons with human evaluations, multiple datasets are needed to ensure reliability. To find an optimal ensemble, a heuristic approach was adopted, involving repeated trial-and-error using numerous low-performing models.
One limitation is the lack of a systematic and efficient method for identifying which weak learners contribute effectively to the ensemble.
Although the BI method itself is based on statistical processing and thus avoids the risk of excessive tuning, the ensemble construction process may still be susceptible to overfitting or over-tuning.

Future research should focus on generalizing the effectiveness of the EBI method through verification using more diverse datasets, investigating systematic selection criteria for effective weak learners, introducing human evaluation for various natural language data, and developing methods to efficiently optimize ensemble structures without excessive trial-and-error.

Overall, this research addresses themes of significant importance for the widespread adoption and development of AI technologies, such as constructing high-performance AI systems in resource-limited situations and effectively utilizing individually low-performing systems. In particular, the combination of SLM ensembles and the EBI method represents a promising direction toward realizing more efficient and robust NLP. By overcoming these challenges and expanding the range of application of these methods, SLM ensembles are expected to develop into more practical technologies in the future.

\appendix
\counterwithin{enumi}{section}
\setcounter{equation}{0}
\section{Samples of Input Data}\label{sec:ABtxt}

This appendix serves to provide examples of the data used in the identity matching experiments, showing how different descriptions of the same person were given in datasets A and B. Two different identification numbers (id\_A and id\_B) are assigned to the two types of datasets (A and B), and experiments are conducted to identify the ids of the same person. Examples include aptitude assessment and behavioral profiling based on purchase data.

\begin{table}[H] 
\centering
\caption{Sample of identified data for job aptitude assessment. Attribute information: id\_B=1, id\_A=25, Type=1, Age=30.}
\label{tb:sample_pro1}
\begin{tabularx}{\textwidth}{|l|X|} \hline
Category & Description \\ \hline
Strength(A) & Highly responsible, perseverant \\ \hline
Weakness(A) & Prone to stress, averse to change \\ \hline
Assessment(A) & The person has a strong sense of responsibility and perseverance to complete assigned roles. They may feel uneasy about adapting to changes, but by clearly defining goals, schedules, and expectations, they can effectively lead team ... \\ \hline
Personnel(B) & As a manager, the employee leads the team and delivers the expected results. To further enhance the overall output of the team, please focus on strategic goal setting, progress management, and member development. It is also ... \\ \hline
\end{tabularx}
\end{table}

\begin{table}[H] 
\centering
\caption{Sample of identified data for purchase behavioral profiling. Attribute information: id\_B=31, id\_A=1, Sex=2, Age=60.}
\label{tb:sample_pro2}
\begin{tabularx}{\textwidth}{|l|X|} \hline
Category & Description  \\ \hline
Style(B) & She values fermented foods and prefers natural foods and nutrient-rich ingredients.  \\ \hline
Persona(B) & She has a strong interest in fermentation and health, especially favoring fermented foods like yogurt and kimchi. She is a loyal customer who makes regular purchases and is interested in new fermentation trends, ... \\ \hline
Style(A) & Her loyalty is still low and she is exploring. She has a very strong health-consciousness and a high interest in beauty and health. \\ \hline
Persona(A) & She has a strong interest in anti-aging and maintaining cognitive abilities, prefers natural and additive-free products. She is also sensitive to new beauty trends and is exploring effective beauty foods. \\ \hline
\end{tabularx}
\end{table}

\counterwithin{enumi}{section}
\setcounter{equation}{0}
\section{Prompts for BI Method}\label{sec:EBpr}

Below are examples of prompts used when estimating $A$ from $B$. Type 1 prompt t1 outputs the pairs of ids deemed identical, while Type 2 prompt t2 outputs judgments along with subjective confidence $s_{ij}$ source data. In this example, the number of ids to match is set to 7. When sending the prompt, individual text data is inserted.

\subsection{Type 1 Prompt: t1 (aptitude assessment)}\label{t1}

\noindent
You are an expert profiler skilled in analyzing people's personalities and psychology.
Based on \#\#Personnel Evaluation Findings of id\_B, infer the profile of the individual 
and compare it with id\_A, which appears to be the same person, from \#\#Aptitude Assessment Findings.
As an expert, select the most likely match according to \#\#Analysis Approach 
and \#\#Execution Method, and output the result in the specified \#\#Output Format. 

\noindent
\#\#Analysis Approach\\
*Emulate human thinking processes and conduct qualitative analysis to draw conclusions.\\
*Directly interpret the data and make intuitive inferences from the context and expressions.\\
*Analyze the individual's behavioral traits, professional abilities, and personal characteristics 
in detail based on the comment from \#\# Personnel Evaluation Findings of id\_B, and estimate the profile.\\
*Compare the inferred profile with the comment from \#\#Aptitude Assessment Findings 
of id\_A and select the candidate id\_A that most closely matches.

\noindent
\#\#Execution Method\\
*Describe the process of selecting the candidate id\_A that most closely matches the inferred profile.\\
*Once the matching candidate id is found, output that id.\\
*Output the matching candidate id according to the specified \#\# Output Format.

\noindent
\#\#Output Format\\
Describe the process of selecting the candidate id\_A that most closely matches the inferred profile.
id\_B:\{id\_B number\}, id\_A:\{matching candidate id\_A number\}

\noindent
\#\#Aptitude Assessment Findings\\
Comments from the assessment test for id\_A. The data is as follows:\\
\{id\_A, Assessment(A)
 [ repeat 7 sample data ]\}

\noindent
\#\#Personnel Evaluation Findings\\
Comments from the personnel evaluation for id\_B. The data is as follows: \\
\{id\_B, Personnel evaluation(B)
 [ repeat target id data ]\}

\noindent
Based on the above requirements, please output the matching id according to the output format.

\subsection{Type 2 Prompt: t2 (aptitude assessment)}\label{t2}

\noindent
You are an excellent profiler. Please find the id\_A that seems to represent the same person as 
inferred from the Personnel evaluation of id\_B, by comparing it to the Assessment test of 
candidate id\_As. Then, arrange these candidates in order of likelihood. Additionally, output 
the certainty level for each match. Consider the following \#\#Guidelines, \#\#Detailed requirements, 
\#\#Evaluation Method for Certainty Level, \#\#Output Format, and \#\#Data Descriptions:

\noindent
\#\#Guidelines\\
*Mimic human thought processes and derive results through qualitative analysis.\\
*Read the content of the data directly and intuitively infer from its context and expressions.\\
*Based on the Personnel evaluation of id\_B, analyze the person's behavioral characteristics, 
professional abilities, and personal traits in detail to infer the persona.\\
*Compare the inferred persona with the Assessment test of id\_A to determine the certainty level of a match.

\noindent
\#\#Detailed Requirements\\
*Describe the inferred persona. Compare the inferred persona with the Assessment test of id\_A to find matching candidates. 
Calculate the certainty level (in percentage).\\
*List the matching candidate id\_As in order of highest certainty level.\\
*Output up to the 7 matching candidate id\_As.\\
*Display the certainty level next to each matching id\_A.\\
*Output the results for all id\_B (7 in total) in the specified \#\#Output Format without omitting any steps.

\noindent
\#\#Evaluation Method for Certainty Level\\
High certainty (e.g., 0.9 - 1.0): A very clear match between both texts.\\
Medium certainty (e.g., 0.5 - 0.8): Some commonalities exist, but it is not a perfect match.\\
Low certainty (e.g., 0.1 - 0.4): Not very confident, but it is a possible match.\\
Very low certainty (e.g., 0.0): Little to no matching points between the texts.

\noindent
\#\#Output Format\\
**id\_B:\{id\_B number\}** \{Description of the inferred persona.\}\\
1. id\_B:\{id\_B number\}, id\_A:\{matching candidate id\_A number\} 
\{certainty level\}\\
2. id\_B:\{id\_B number\}, id\_A:\{matching candidate id\_A number\} 
\{certainty level\}\\
(Omitted)

\noindent
\#\#Data Descriptions\\
Below are the two CSV datasets to be used for the analysis.\\
* **CSV data describing id\_A:** The data includes id\_A, Assessment test. \\
The data is as follows in the \{\}.\\
\{id\_A, Assessment(A) [repeat 7 sample data ]\}\\
* **CSV data describing id\_B:** The data includes id\_B, Personnel evaluation. \\
The data is as follows in the \{\}.\\
\{id\_B, Personnel evaluation(B) [ repeat 7 sample data ] \}

\noindent
Based on the above requirements, please output their matching candidate id\_A and the certainty levels for all id\_B, following \#\#Output Format, up to the 7 matching candidate. 
No code is needed.

\setcounter{equation}{0}
\section{Sequential Thinking Prompting}\label{sec:Spr}

Using prof1e (English/aptitude assessment) as an example, the details of sequential prompting are described. The prompt consists of four parts: specifying the profile analysis method (S1), processing sequentially by comparing one person at a time in a step-by-step manner (S2), activating recursive processing when the remaining number reaches a specified value (default=2) (S3), and performing duplication checking and review through S4.

\noindent
{\bf 1. System prompt S1}\\
You are an exceptional profiler specializing in identity matching. Execute the judgment step by step.
For each `id\_B` in the B file, estimate which `id\_A` in the A file it corresponds to.
Follow the \{\#Analysis Method\} described below and make judgments step by step 
with the highest possible accuracy.

\noindent
\# Description of Input Files\\
`id\_A` and `id\_B` are both employee IDs, but the numbers have no correlation.\\
Each `id\_A` corresponds to exactly one `id\_B`.\\
- **A File**: Contains employee ID (`id\_A`), age, type, strengths, weaknesses, and aptitude test results.\\
- **B File**: Contains employee ID (`id\_B`), age, type, and HR comments.

\noindent
\# Analysis Method\\
- **Analysis of Each `id\_B`**: Carefully analyze behavioral traits and personal characteristics 
inferred from the age, type, and HR comments to create a detailed profile of the individual. \\
- **Comparison of Similar Candidates**: Identify `id\_A` candidates with the same age and type 
as `id\_B` and evaluate which `id\_A` profile most closely matches the characteristics of `id\_B`. 
Justify the conclusion by comparing the information in the A file (age, strengths, weaknesses, aptitude test results). 

\noindent
{\bf 2. Sequential processing S2}\\
Please evaluate `id\_B=\{id\_b\}`. Provide your answer in JSON format as follows:\\
 `\{\{"thought": str, "id\_A": int\}\}`. \\
In `thought`, record the step-by-step comparison and reasoning between `id\_B` and `id\_A`. 
For `id\_A`, enter the ID number of the individual who is most similar. \\
\#\# B File \\
\{row\_b\} \\
$[$ Conditional Branching: the prompt will be changed as follows depending on k. $]$ \\
if the counter k= n, """ \\
The matching candidates are as follows: `id\_A=\{ids\_a\}`. \\
\#\# A File \\
\{rows\_a\} """ \\
else"""The target for evaluation is id\_A=\{ids\_a\}. The information was provided earlier."""

\noindent
{\bf 3. Recursion S3}\\
Please evaluate `id\_B=\{ids\_b\}`. The evaluation targets will be `id\_A=\{ids\_a\}`. \\
If there is any inconsistency or duplication in the evaluation results, you can reselect 
from the initial candidate set S0=id\_A\{\{\{orig\_ids\_a\}\}\}. \\
If there are duplicate evaluation results, determine which is more plausible.
Additionally, review previous evaluation results, identify any `id\_B` that require 
corrections in their linkage to `id\_A`, and make adjustments if necessary.\\
\noindent
Provide your response in a written format. First, describe the step-by-step comparison
and reasoning process. Next, document any revisions to the evaluation results. \\
Finally, output the confirmed pairs of `id\_B` and `id\_A`.
For the ID pairs, include all previous `id\_B=\{old\_ids\_b\}`.
\#\# B File
\{rows\_b\}

\noindent
{\bf 4. Conflict resolution S4}\\
Using the output pairs of `id\_A` and `id\_B`, check for duplicates or unresolved IDs by following these steps:\\
\#\# Steps\\
- Within the `<thinking>` tag, document the review process for the evaluation results.
Include considerations such as:\\
- Ensuring no duplicate `id\_A` is assigned to multiple `id\_B`.\\
- Resolving any unresolved IDs by determining the most plausible pairs.\\
- Reviewing the overall results to replace any pairs with more plausible ones.\\
- Within the `<result>` tag, record the revised pairs of IDs. If no revisions are necessary,
document the original pairs as they are.\\
- Within the `<reflection>` tag, reflect on whether there are any duplicates 
   or unresolved IDs in the revised results and record the status.\\
- Within the `<count>` tag, indicate the number of additional revisions required. 
   If no further revisions are needed, set the value in `<count>` to 0.

\counterwithin{enumi}{section}
\setcounter{equation}{0}
\section{Tables of Results}\label{sec:alldata}

This appendix lists the results for each system divided by dataset. Table items for single BI systems include the system number, the name of generative AI model, prompt used to generate observation matrix $c_{ji}$ (type and call count), prompt used to generate weight matrix $s_{ij}$ (type and call count), number of correct answers $n_c$, Lift, and Reach.
The asterisk on t1$\ast$ refers to prompts that require description of the judgment process.
The prime on t2$'$ denotes instances where groq:llama3-70b-8192 was used for generating $s_{ij}$.

For the Tables of EBI results, the item "components" lists the set of single BI system numbers comprising the ensemble, and "weights" shows their respective weights.

\subsection{prof1j: Japanese Aptitude Assessment}\label{sec:prof1j}

In this case, the baseline values for (Lift, Reach) are $(H,G)=(19,22)$.
Table~\ref{tb:single_pro1} shows the results of BI systems. BI systems 37 and 42 achieved results equivalent to system 76 (llama3.1-70b-versatile) with a larger size of model. They also surpassed systems 64 and 66 (gpt-4o-mini-2024-07-18).
Table~\ref{tb:complex_pro1} shows the results of EBI (ensemble BI systems).
All of them demonstrate that by combining BI systems (66, 40, 12, 13) with negative Lift, the Lift values turned positive. 
The Acc of EBI systems 50, 81, and 83 exceeded that of single systems.

\begin{table}[H] 
\centering
\caption{Results of single BI systems for prof1\_j (Japanese Aptitude Assessment). }
\label{tb:single_pro1}
\begin{tabular}{|c|l|c|c|c|c|c|} \hline
   system & model & $c_{ji}$ & $s_{ij}$ & $n_c$ & Lift & Reach \\ \hline
   37 & gemma2-9b-it & t1$\ast$-100 & t2$'$-10 & 23 &21.1\% &104.5\%  \\ \hline
   42 & llama3-8b-8192 & t1$\ast$-100 & t1$\ast$-100 & 23 &21.1\% &104.5\%  \\ \hline
   76 & llama3.1-70b-versatile & t1$\ast$-100 & t2-10 & 23 &21.1\% &104.5\%  \\ \hline
   64 & gpt-4o-mini-2024-07-18 & t2-10 & t2-10 & 21 &10.5\% &95.5\%  \\ \hline
   43 & llama3-8b-8192 & t1$\ast$-100 & t2$'$-10 & 21 &10.5\% &95.5\%  \\ \hline
   25 & gemma2-9b-it & t1$\ast$-100 & t2$'$-10 & 20 &5.3\% &90.9\%  \\ \hline
   28 & llama3-8b-8192 & t1$\ast$-100 & t1$\ast$-100 & 20 &5.3\% &90.9\% \\ \hline
   46 & llama3-70b-8192 & t1$\ast$-100 & t2-10 & 20 &5.3\% &90.9\% \\ \hline
   66 & gpt-4o-mini-2024-07-18 & t1$\ast$-100 & t2-10 & 18 &-5.3\% &81.8\% \\ \hline
   40 & mixtral-8x7b-32768 & t1$\ast$-100 & t2$'$-10 & 18 &-5.3\% &81.8\% \\ \hline
   12 & mixtral-8x7b-32768 & t1-500 & t1-500 & 18 &-5.3\% &81.8\% \\ \hline
   13 & llama3-70b-8192 & t1$\ast$-500 & t1$\ast$-500 & 17 &-10.5\% &77.3\% \\ \hline
\end{tabular}
\end{table}

\begin{table}[H] 
\centering
\caption{Results of EBI (ensemble systems) for prof1\_j* (Japanese Aptitude Assessment), limited to top-performing systems (Lift$\geq0$).}
\label{tb:complex_pro1}
\begin{tabular}{|c|c|c|c|c|c|} \hline
system & components & weights & $n_c$ & Lift &Reach \\ \hline
83,81 & \{37,40,43,46\} & [1,1,1,1],[1,1,2,3]& 26 &36.8\% &118.2\% \\ \hline
50 &  \{12,13,25,28,37,40,43,46\} & [3,2,1,1,1,1,2,3] & 25 &31.6\% &113.6\% \\ \hline
55 &  \{12,13,25,28,37,40,43,46\} & [3,2,1,1,5,1,2,3] & 23 &21.1\% &104.5\% \\ \hline
78 &  \{37,43,45,66,76\} & [30,3,1,1,10] & 23 &21.1\% &104.5\% \\ \hline
71 &\{37,43,66\} & [1,1,1]& 22 &15.8\% &100.0\% \\ \hline
82,84 &  \{37,40,43,46,66,76\} & [1,1,2,3,1,1],[1,1,1,1,1,1] & 20 &5.3\% &90.9\% \\ \hline
85 &  \{37,40,42,46,64,59\} & [1,1,1,1,1,1] & 19 &0.0\% &86.4\% \\ \hline 
\end{tabular}
\end{table}

\subsection{prof2j: Japanese Purchase Data}\label{sec:prof2j}

In this case, the baseline values for (Lift, Reach) are $(H,G)=(13,20)$.
Table~\ref{tb:single_pro2} shows the results of BI systems,
and Table~\ref{tb:complex_pro2} shows the results of EBI (ensemble BI systems).
They are composed primarily of BI systems (37, 46, 64, 76) that achieved positive Lift in prof1j.
Combining systems with negative or zero Lift (40, 42, 43) resulted in positive Lift values.

The top-performing EBI system 82 includes the system 76 of size 70b, but it surpasses the Lift value of system 76 alone.
EBI system 89 \{37, 40, 43\}, composed of systems without 70b model, achieves accuracy equivalent to system 37 alone while maintaining positive Lift. These results suggest that using multiple SLMs together, rather than relying heavily on systems with large sizes like 70b, can provide diversity and robustness.

\begin{table}[H] 
\centering
\caption{Results of single BI systems for prof2\_j (Japanese Purchase Data).}
\label{tb:single_pro2}
\begin{tabular}{|c|l|c|c|c|c|c|} \hline
system & model & $c_{ji}$ & $s_{ij}$ & $n_c$ & Lift & Reach \\ \hline
59 & cerebras:llama3.1-70b-versatile & t2-10 & t2-10 & 21 &61.5\% &105\% \\ \hline   
46 & llama3-70b-8192 & t1$\ast$-100 & t2-10 & 17 &30.8\% &85\% \\ \hline
64 & gpt-4o-mini-2024-07-18 & t2-10 & t2-10 & 16 &23.1\% &80\% \\ \hline
65 & gpt-4o-mini-2024-07-18 & t1$\ast$-100 & t1$\ast$-100 & 16 &23.1\% &80\% \\ \hline
76 & cerebras:llama3.1-70b-versatile & t1$\ast$-100 & t2-10 & 16 &23.1\% &80\% \\ \hline
66 & gpt-4o-mini-2024-07-18 & t1$\ast$-100 & t2-10 & 15 &15.4\% &75\% \\ \hline
37 & gemma2-9b-it & t1$\ast$-100 & t2$'$-10 & 15 &15.4\% &75\% \\ \hline
40 & mixtral-8x7b-32768 & t1$\ast$-100 & t2$'$-10 & 13 &0.0\% &65\% \\ \hline
45 & llama3-70b-8192 & t1$\ast$-100 & t1$\ast$-100 & 12 &-7.7\% &60\% \\ \hline
42 & llama3-8b-8192 & t1$\ast$-100 & t1$\ast$-100 & 12 &-7.7\% &60\% \\ \hline
43 & llama3-8b-8192 & t1$\ast$-100 & t2$'$-10 & 11 &-15.4\% &55\% \\ \hline
 \end{tabular}
\end{table}

\begin{table}[H] 
\centering
\caption{Results of EBI (ensemble systems) for prof2\_j* (Japanese Purchase Data), limited to top-performing systems (Lift$\geq0$).}
\label{tb:complex_pro2}
\begin{tabular}{|c|c|c|c|c|c|} \hline
 system & components & weights & $n_c$ & Lift & Reach\\ \hline
 82 &  \{37,40,43,46,66,76\} & [1,1,2,3,1,1] & 18 &38.5\% &90\% \\ \hline
 83 & \{37,40,43,46\} & [1,1,1,1]& 17 &30.8\% &85\% \\ \hline
 84 &  \{37,40,43,46,66,76\} & [1,1,1,1,1,1] & 16 &23.1\% &80\% \\ \hline
 85 &  \{37,40,42,46,64,59\} & [1,1,1,1,1,1] & 16 &23.1\% &80\% \\ \hline
 81,91 &  \{37,40,43,46\} & [1,1,2,3], [1,1,1,1] & 15 &15.4\% &75\% \\ \hline
 89 &  \{37,40,43\} & [1,1,1] & 15 &15.4\% &75\% \\ \hline
90 &  \{37,40,43\} & [1,2,1] & 14 &7.7\% &70\%  \\ \hline
92,93 &  \{37,40,43,46\} & [1,2,1,2], [1,2,1,3] & 14 &7.7\% &70\% \\ \hline
 86 &  \{37,40,59\} & [1,1,1] & 14 &7.7\% &70\% \\ \hline
 74 &  \{37,64\} & [1,1] & 13 &0.0\% &65\% \\ \hline
 77,78 & \{37,43,45,66,76\} & [1,1,1,1,1],[30,3,1,1,10]& 12 &-7.7\% &60\% \\ \hline
\end{tabular}
\end{table}

\subsection{prof1e: English Aptitude Assessment}\label{sec:prof1e}
For prof1e, the baseline reference values are
$(H,G)=(21.2,28)$, where $H=G*\gamma$, $\gamma=0.757$.
Table~\ref{tb:single_pro1e} shows the results of BI systems. System 40 (mixtral-8x7b-32768) achieved results equivalent to system 66 (gpt-4o-mini-2024-07-18) and surpassed system 76 (llama3.1-70b-versatile).

Table~\ref{tb:complex_pro1e} shows the results of EBI. 
System 93 \{37, 40, 43, 76\} surpassed all single systems, and while its Reach against GPT-4o did not reach 100\%, it achieved a notably high 96\%. It is interesting that adding BI systems 37, 43, and 76, which had low Reach individually, improved the overall Reach. Additionally, EBI systems 89 and 90 \{37, 40, 43\} achieved Reach equivalent to the highest single system while using weak systems 37 and 43 (gemma2-9b-it, llama3-8b-8192) to ensure diversity and serve as sources of robustness.

\begin{table}[H] 
\centering
\caption{Results of single BI systems for prof1\_e (English Aptitude Assessment).}
\label{tb:single_pro1e}
\begin{tabular}{|c|l|c|c|c|c|c|} \hline
 system & model & $c_{ji}$ & $s_{ij}$ & $n_c$ & Lift & Reach \\ \hline
   66 & gpt-4o-mini-2024-07-18 & t1$\ast$-100 & t2-10 & 25 &17.9\% &89.3\% \\ \hline
   40 & mixtral-8x7b-32768 & t1$\ast$-100 & t2$'$-10 & 25 &17.9\% &89.3\% \\ \hline
   76 & cerebras:llama3.1-70b-versatile & t1$\ast$-100 & t2-10 & 22 &3.8\% &78.6\% \\ \hline
   46 & llama3-70b-8192 & t1$\ast$-100 & t2-10 & 21 &-0.9\% &75.0\% \\ \hline
   37 & gemma2-9b-it & t1$\ast$-100 & t2$'$-10 & 20 &-5.7\% &71.4\% \\ \hline
   43 & llama3-8b-8192 & t1$\ast$-100 & t2$'$-10 & 20 &-5.7\% &71.4\% \\ \hline
\end{tabular}
\end{table}

\begin{table}[H] 
\centering
\caption{Results of EBI (ensemble systems) for prof1\_e* (English Aptitude Assessment).}
\label{tb:complex_pro1e}
\begin{tabular}{|c|c|c|c|c|c|} \hline
 system & components & weights & $n_c$ &Lift & Reach \\ \hline
93 & \{37,40,43,76\} & [1,2,1,3] & 27 &27.4\% &96.4\% \\ \hline
89,90 & \{37,40,43\} & [1,1,1],[1,2,1] & 25 &17.9\% &89.3\% \\ \hline
91 & \{37,40,43,66\} & [1,1,1,1]& 24 &13.2\% &85.7\% \\ \hline
83 &  \{37,40,43,46\} & [1,1,1,1] & 24 &13.2\% &85.7\% \\ \hline
92 & \{37,40,43,66\} & [1,2,1,2] & 23 &8.5\% &82.1\% \\ \hline
84 &  \{37,40,43,46,66,76\} & [1,1,1,1,1,1] & 22 &3.8\% &78.6\% \\ \hline
81 &  \{37,40,43,46\} & [1,1,2,3] & 22 &3.8\% &78.6\% \\ \hline
71 &\{37,43,66\} & [1,1,1] & 21 &-0.9\% &75.0\% \\ \hline
82 &  \{37,40,43,46,66,76\} & [1,1,2,3,1,1] & 19 &-10.4\% &67.9\% \\ \hline
\end{tabular}
\end{table}

\subsection{prof2e: English Purchase Data}\label{sec:prof2e}
For prof2e, the baseline reference values are
$(H,G)=(17.4,23)$, where $H=G*\gamma$, $\gamma=0.757$.
Table~\ref{tb:single_pro2e} shows the results of BI systems.
System 37 (gemma2-9b-it) achieved Reach$>$100\%, surpassing GPT-4o.
System 66 (gpt-4o-mini) similarly exceeded GPT-4o.

Table~\ref{tb:complex_pro2e} shows the results of EBI. Systems 95 and 96, which synthesize all six single systems mentioned above, rank as the top 2. Systems excluding system 94 (llama3.3-70b) such as 92 and 93 also achieved Reach$>$100\%, surpassing GPT-4o. These systems include weak systems 40 and 43, which can be considered sources of robustness.

\begin{table}[H] 
\centering
\caption{Results of single BI systems for prof2\_e (English Purchase Data).}
\label{tb:single_pro2e}
\begin{tabular}{|c|l|c|c|c|c|c|} \hline
 system & model & $c_{ji}$ & $s_{ij}$ & $n_c$ &Lift & Reach \\ \hline
   94 & cerebras:llama3.3-70b & t1$\ast$-100 & t2-10 & 30 &72.4\% &130.4\% \\ \hline
   46 & llama3-70b-8192 & t1$\ast$-100 & t2-10 & 29 &66.7\% &126.1\% \\ \hline
   37 & gemma2-9b-it & t1$\ast$-100 & t2$'$-10 & 25 &43.7\% &108.7\% \\ \hline
   66 & gpt-4o-mini-2024-07-18 & t1$\ast$-100 & t2-10 & 24 &37.9\% &104.3\% \\ \hline
   40 & mixtral-8x7b-32768 & t1$\ast$-100 & t2$'$-10 & 17 &-2.3\% &73.9\% \\ \hline
   43 & llama3-8b-8192 & t1$\ast$-100 & t2$'$-10 & 17 &-2.3\% &73.9\% \\ \hline
\end{tabular}
\end{table}

\begin{table}[H] 
\centering
\caption{Results of EBI (ensemble systems) for prof2\_e* (English Purchase Data).}
\label{tb:complex_pro2e}
\begin{tabular}{|c|c|c|c|c|c|} \hline
 system & components & weights & $n_c$ &Lift & Reach \\ \hline
96 & \{37,40,43,46,66,94\} & [1,1,1,1,1,1] & 31 &78.2\% &134.8\% \\ \hline
95 & \{37,40,43,46,66,94\} & [1,1,2,3,1,1] & 29 &66.7\% &126.1\% \\ \hline
92,93 & \{37,40,43,66\} & [1,2,1,2],[1,2,1,3] & 28 &60.9\% &121.7\% \\ \hline
81,83 &  \{37,40,43,46\} & [1,1,2,3],[1,1,1,1] & 26 &49.4\% &113.0\% \\ \hline
91 & \{37,40,43,66\} & [1,1,1,1]& 26 &49.4\% &113.0\% \\ \hline
71 &\{37,43,66\} & [1,1,1] & 23 &32.2\% &100.0\% \\ \hline
89,90 & \{37,40,43\} & [1,1,1],[1,2,1] & 23 &32.2\% &100.0\% \\ \hline
\end{tabular}
\end{table}

\newcommand{\doi}[2]{\,\href{#1}{#2}\,}  



\begin{thebibliography}{99}

\bibitem{ref1}
A. Rodman, T. A. Buckley, A. K. Manrai, and D. J. Morgan, "Artificial Intelligence vs Clinician Performance in Estimating Probabilities of Diagnoses Before and After Testing", JAMA Network Open. 2023;6(12):e2347075
\bibitem{ref2}
P.-H. Chen, J.-J. Jung, Y. Kim, M. Lee et.al, "AI-assisted clinical summary and treatment planning for cancer care: A comparative study of human vs. AI-based approaches.(AI-Assisted Planning and Summarization)", Journal of Clinical Oncology Volume 42, 1523-1523(2024).
\bibitem{ref3}
A. Rodman, T. A. Buckley, A. K. Manrai, and D. J. Morgan, "Accuracy of a Generative Artificial Intelligence Model in a Complex Diagnostic Challenge", JAMA Network Open. 2023;6(12):e2347075
\bibitem{ref4}
E. Goh, R. Gallo, J. Hom, E. Strong, Y. Weng, et.al,  "Large Language Model Influence on Diagnostic Reasoning", JAMA Network Open. 2024;7(10):e2440969
\bibitem{ref5}
Kanjee Z, Crowe B, Rodman A. Accuracy of a Generative Artificial Intelligence Model in a Complex Diagnostic Challenge. JAMA. 2023;330(1):78-80
\bibitem{ref6}
N. Bian, X. Han, L. Sun, H. Lin, Y. Lu, B. He, S. Jiang, and B. Dong. 2024. ChatGPT Is a Knowledgeable but Inexperienced Solver: An Investigation of Commonsense Problem in Large Language Models. In Proceedings of the 2024 Joint International Conference on Computational Linguistics, Language Resources and Evaluation (LREC-COLING 2024), pages 3098–3110, Torino, Italia. ELRA and ICCL.
\bibitem{ref7}
C. Si, D. Yang, T. Hashimoto, "Can LLMs Generate Novel Research Ideas? A Large-Scale Human Study with 100+ NLP Researchers", International Conference on Learning Representations (ICLR) 2025.
https://doi.org/10.48550/arXiv.2409.04109
\bibitem{ref8}
Z. Cui, N. Li, H. Zhou, "Can AI Replace Human Subjects? A Large-Scale Replication of Psychological Experiments with LLMs", 
Available at SSRN: https://dx.doi.org/10.2139/ssrn.4940173
\bibitem{ref9}
Q. Mei, Y. Xie, W. Yuan, and M. O. Jackson, "A Turing Test of Whether AI Chatbots Are Behaviorally Similar to Humans", Proceedings of the National Academy of Sciences, Vol. 121, No. 9, 2024
\bibitem{ref10}
Tartaraj, A., and Trebicka, B. (2023). "Optimal grading scales for enhancing product evaluation by infomediaries", Interdisciplinary Journal of Research and Development, 10(1), 59–66.
\bibitem{ref11}
Tholen, G. (2024). "Matching candidates to culture: How assessments of organisational fit shape the hiring process", Work, Employment and Society, 38(3), 705–722.
\bibitem{ref12}
X. Zhai, M. Nyaaba, W. Ma, "Can generative AI and ChatGPT outperform humans on cognitive-demanding problem-solving tasks in science?", Epistemic Insight \& Artificial Intelligence(2024).
\bibitem{ref13}
T. Szanda\l a, "ChatGPT vs human expertise in the context of IT recruitment", Expert Systems with Applications, Volume 264, 10 March 2025, 125868
\bibitem{ref14}
J. Qiu, D. Guo, P. Natalie, P. Noelle, L. Cheri, T. R. Henry, "Ensemble of Large Language Models for Curated Labeling and Rating of Free-text Data", 	arXiv:2501.08413 [cs.CL], https://doi.org/10.48550/arXiv.2501.08413

\bibitem{ref15}
C. Zhang, L. Liu, J. Wang, C. Wang, X. Sun, H. Wang, M. Ca, "PREFER: Prompt Ensemble Learning via Feedback-Reflect-Refine", Proceedings of the AAAI Conference on Artificial Intelligence, 38(17)

\bibitem{ref16}
Jiang, D., Ren, X., Lin, B.Y.(2023). "LLM-Blender: Ensembling Large Language Mod-els with Pairwise Ranking and Generative Fusion",  In Proceedings of the 61st Annual Meeting of the Association for Computational Linguistics (Volume 1: Long Papers), pages 14165–14178, Toronto, Canada. Association for Computational Linguistics.
\bibitem{ref17}
Lu, K., Yuan, H., Lin, R., Lin, J., Yuan, Z., Zhou, C., Zhou, J.(2024). "Routing to the Expert: Efficient Reward-guided Ensemble of Large Language Models", In Proceedings of the 2024 Conference of the North American Chapter of the Association for Computational Linguistics: Human Language Technologies (Volume 1: Long Papers), pages 1964–1974, Mexico City, Mexico. Association for Computational Linguistics.
\bibitem{ref18}
Shinn, N., Cassano, F., Gopinath, A., Narasimhan, K., and Yao, S. (2023), "Reflexion: Language Agents with Verbal Reinforcement Learning", Advances in Neural Information Processing Systems, 36.
\bibitem{ref19}
S. Dhuliawala, M. Komeili, J. Xu, R. Raileanu, X. Li, A. Celikyilmaz, and J. Weston. 2024. "Chain-of-Verification Reduces Hallucination in Large Language Models", In Findings of the Association for Computational Linguistics: ACL 2024, pages 3563–3578, Bangkok, Thailand. Association for Computational Linguistics.


\end{thebibliography}
\end{document}